\documentclass[conference]{IEEEtran}
\usepackage{cite}
\usepackage{amsmath,amssymb,amsfonts}
\usepackage{algorithmic}
\usepackage{graphicx}
\usepackage{textcomp}
\usepackage{xcolor}
\usepackage{booktabs}
\usepackage{hyperref}
\usepackage{balance}
\usepackage{multirow}

\begin{document}

\title{TurboVec: A Case Study in Cost-Efficient Private Retrieval for Enterprise RAG via Codebook-Oblivious Quantization}

\author{\IEEEauthorblockN{Navnit Shukla}
\IEEEauthorblockA{Principal AI Architect \\
Snowflake Inc.\\
Tustin, California, USA\\
\url{https://orcid.org/0009-0005-8801-3344}}
\and
\IEEEauthorblockN{Kamal Pandey}
\IEEEauthorblockA{Enterprise Applied AI\\
Rivian Automotive Inc.\\
Irvine, United States of America\\
\url{https://orcid.org/0009-0009-1669-4450}}
\and
\IEEEauthorblockN{Omshankar Tiwari}
\IEEEauthorblockA{Applied AI Technical Lead \\
Google Deepmind\\
Sunnyvale, California, USA\\
\url{https://orcid.org/0009-0001-2247-0383}}}

\maketitle

\begin{abstract}
Retrieval-Augmented Generation (RAG) systems increasingly power enterprise LLM applications, yet their vector retrieval layer introduces two underexplored challenges: (1) trained codebook quantizers may expose corpus statistics during index construction, creating a potential leakage channel in multi-tenant deployments, and (2) post-hoc filtering for tenant isolation degrades recall on selective queries. We study TurboVec, an open-source vector index built on TurboQuant~\cite{turboquant2026}---a codebook-oblivious scalar quantizer that derives quantization boundaries analytically from known distributional properties of high-dimensional, L2-normalized embeddings, requiring no corpus-dependent training. On the public DBpedia OpenAI embeddings benchmark (d=1536, 100K--999K vectors), TurboQuant 4-bit consistently outperforms trained FAISS Product Quantization at the same 4-bit budget by 8.5--8.9 percentage points in Recall@5 across all scales tested, including near-million-scale evaluation. Compared to HNSW (R@5=0.991) and IVF-PQ (R@5=0.840), TurboQuant occupies a distinct design point: higher recall than IVF-PQ without any training, at 4--8$\times$ less memory than HNSW. In a case study deployment on Snowpark Container Services, TurboVec achieves 11ms median query latency at 100K vectors versus approximately 707ms for a warehouse brute-force scan. Kernel-level multi-tenant allowlist filtering maintains 0.86--0.93 Recall@10 across 10--1000 tenant workloads versus 0.09--0.19 for a simple over-fetch post-filter baseline. We further evaluate codebook-based membership inference under a narrow threat model, finding that TurboVec's codebook-oblivious design reduces the signal of this specific attack class to near-random (50.0\% accuracy) versus 57.3\% for PQ-based codebooks. We acknowledge important limitations: results are on a single dataset and embedding model, the HNSW comparison uses uncompressed FP32 vectors (a compressed HNSW variant may narrow the memory gap), and the privacy evaluation is on synthetic data only.on is not included, and our privacy evaluation covers only codebook-based attacks, not access-pattern or query-level leakage.
\end{abstract}


\section{Introduction}
\label{sec:intro}

Retrieval-Augmented Generation (RAG) has emerged as a dominant architecture for grounding LLM outputs in factual knowledge~\cite{rag_survey_2025}. At the core of every RAG system lies a vector index servicing similarity queries over document embeddings. As enterprises deploy multi-tenant RAG systems on shared infrastructure, two practical challenges arise.

\textbf{Codebook leakage in index construction.} Production quantization methods---Product Quantization (PQ)~\cite{faiss}, Optimized PQ, and learned quantizers---require a training phase on representative data to construct codebooks. In multi-tenant deployments, a trained codebook may encode corpus statistics that an adversary could exploit. While cryptographic privacy-preserving ANN methods~\cite{pacmann,panther,pp_anns} address stronger threat models (hiding queries or databases from the server), they carry substantial overhead. A pragmatic intermediate design point is \emph{codebook-oblivious} quantization: a scheme whose codebook structure is independent of the indexed corpus, eliminating this specific leakage surface at zero protocol overhead.

\textbf{Filtered search efficiency.} Enterprise RAG requires tenant-isolated retrieval. Post-filtering---searching the full index then dropping non-tenant results---degrades recall on selective queries and wastes compute. Kernel-level filtering operates at the SIMD block level, short-circuiting irrelevant blocks before scoring.

TurboQuant~\cite{turboquant2026} (ICLR 2026) provides an analytically-derived scalar quantizer whose boundaries are precomputed from the known distribution of coordinates in high-dimensional, L2-normalized vectors after random rotation---requiring no corpus-dependent training. We evaluate TurboVec, an open-source Rust implementation of TurboQuant, on a single public embedding benchmark (DBpedia entities, d=1536) and one representative cloud deployment. We do not claim general coverage of all enterprise RAG workloads; our focus is the quantization quality-vs-memory tradeoff and codebook-leakage properties in this specific setting.

Our contributions, subject to the scope limitations described in Section~\ref{sec:limitations}, are:

\begin{enumerate}
    \item \textbf{Compression quality scaling study} (Section~\ref{sec:scaling}): TurboQuant 4-bit outperforms trained FAISS PQ 4-bit in Recall@5 across 100K--999K vectors (d=1536), and we compare against HNSW-Flat and IVF-PQ to position TurboQuant within the recall-memory-latency tradeoff space (single dataset; see Limitation~1).
    \item \textbf{Case study deployment} (Section~\ref{sec:eval}): Deployment on Snowpark Container Services (CPU-based flat-scan index) at 100K vectors with measurements of latency, memory, and estimated cost (CPU-only; see Limitation~2).
    \item \textbf{Multi-tenant filtered search} (Section~\ref{sec:eval_filter}): Comparison of kernel-level allowlist filtering versus a simple over-fetch post-filter baseline on synthetic uniform tenant partitions; we do not claim dominance over more sophisticated filtered ANN designs (see Limitation~7).
    \item \textbf{Codebook-based membership inference evaluation} (Section~\ref{sec:privacy}): Under a narrow threat model and a specific quantization-error attack on synthetic d=256 data, TurboVec's codebook-oblivious design reduces attack accuracy to near-random versus PQ codebooks (see Limitation~3).
\end{enumerate}

\begin{figure*}[t]
\centering
\includegraphics[width=\textwidth]{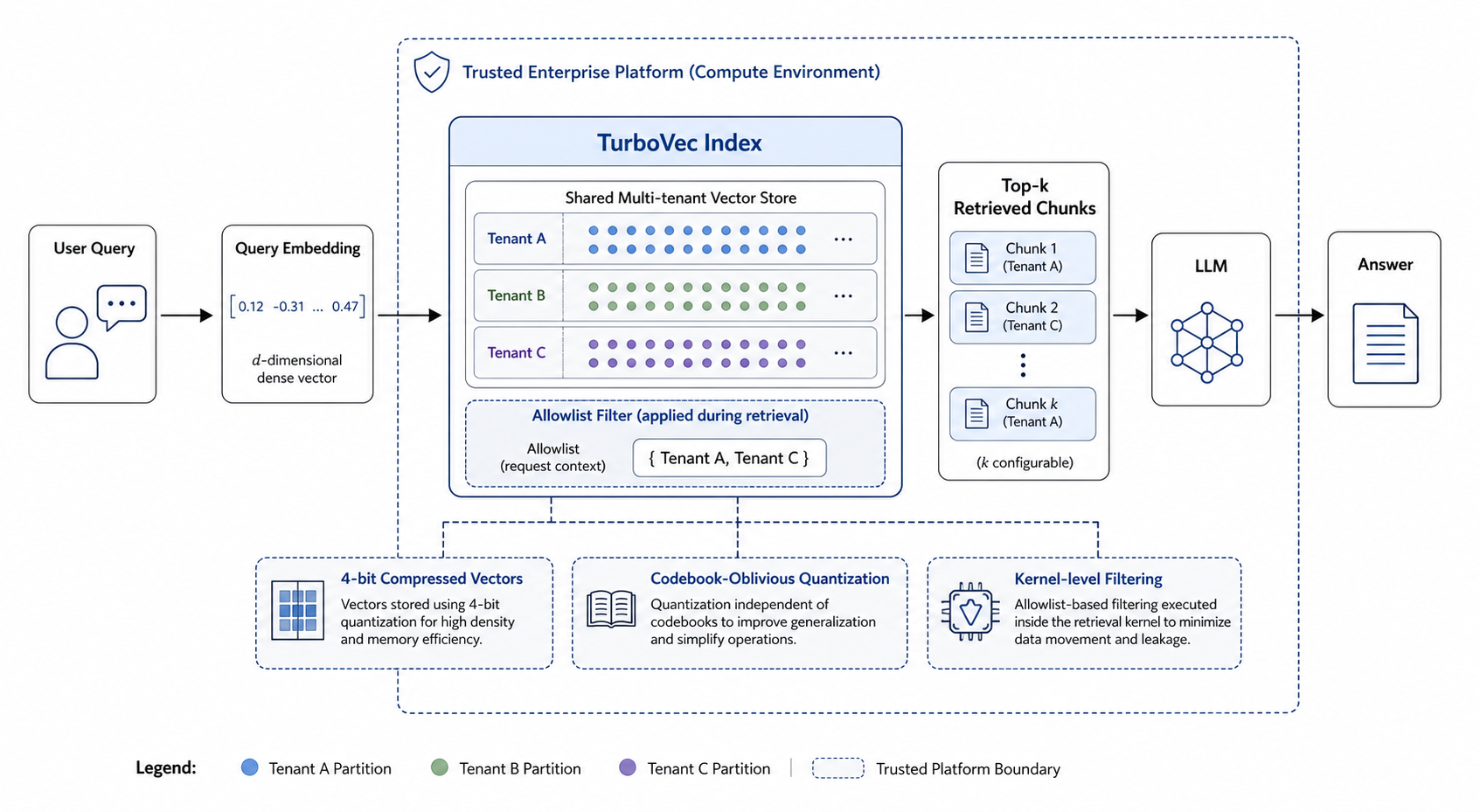}
\caption{TurboVec enterprise RAG pipeline. User queries are embedded, matched against a shared multi-tenant vector store using codebook-oblivious 4-bit quantization, and filtered at the kernel level via per-request allowlists before retrieval results are passed to the LLM.}
\label{fig:rag_pipeline}
\end{figure*}

\section{Background}
\label{sec:background}

\subsection{Terminology: Codebook-Oblivious vs.\ Data-Oblivious}

We use the term \emph{codebook-oblivious} to describe TurboQuant: the codebook boundaries are derived from an analytical distributional model (Beta/Gaussian marginals after random rotation) rather than from the indexed corpus. We deliberately avoid the term \emph{data-oblivious} in its stronger algorithmic sense (ORAM-style obliviousness of memory access patterns~\cite{oram}), which TurboVec does not provide. We note that the optional TQ+ calibration step (Section~\ref{sec:tq_plus}) introduces limited aggregate data dependence, which we analyze separately.

\subsection{TurboQuant Algorithm}

TurboQuant~\cite{turboquant2026} compresses L2-normalized vectors through the following pipeline:
\begin{enumerate}
    \item \textbf{Normalize:} Strip the $\ell_2$ norm, storing it as a scalar.
    \item \textbf{Random rotation:} Multiply by a fixed random orthogonal matrix $R$. After rotation, each coordinate approximately follows a marginal distribution converging to $\mathcal{N}(0, 1/d)$ as $d \to \infty$.
    \item \textbf{Lloyd-Max scalar quantization:} Optimal boundaries are precomputed from the known Beta/$\mathcal{N}(0, 1/d)$ marginal---no data-dependent training.
    \item \textbf{Bit-packing:} Each coordinate becomes a small integer (4 buckets at 2-bit, 16 at 4-bit). A 1536-dim vector compresses from 6,144 bytes to 768 bytes (4-bit).
    \item \textbf{Length-renormalized scoring:} A per-vector scalar corrects inner-product underestimation from quantization.
\end{enumerate}

\subsubsection{TQ+ Calibration and Data Dependence}
\label{sec:tq_plus}

An optional TQ+ step fits per-coordinate shift and scale parameters from the first batch of indexed vectors to align empirical coordinate quantiles with the canonical marginal. This step introduces bounded aggregate data dependence: only per-coordinate first and second moments are retained, not individual vectors. While this departs from strict codebook-obliviousness, the leakage is qualitatively weaker than PQ's $k$-means centroids, which encode cluster structure of the full training corpus. We analyze this distinction empirically in Section~\ref{sec:privacy_tqplus} and leave a formal leakage bound for future work.

\begin{figure}[t]
\centering
\includegraphics[width=\columnwidth]{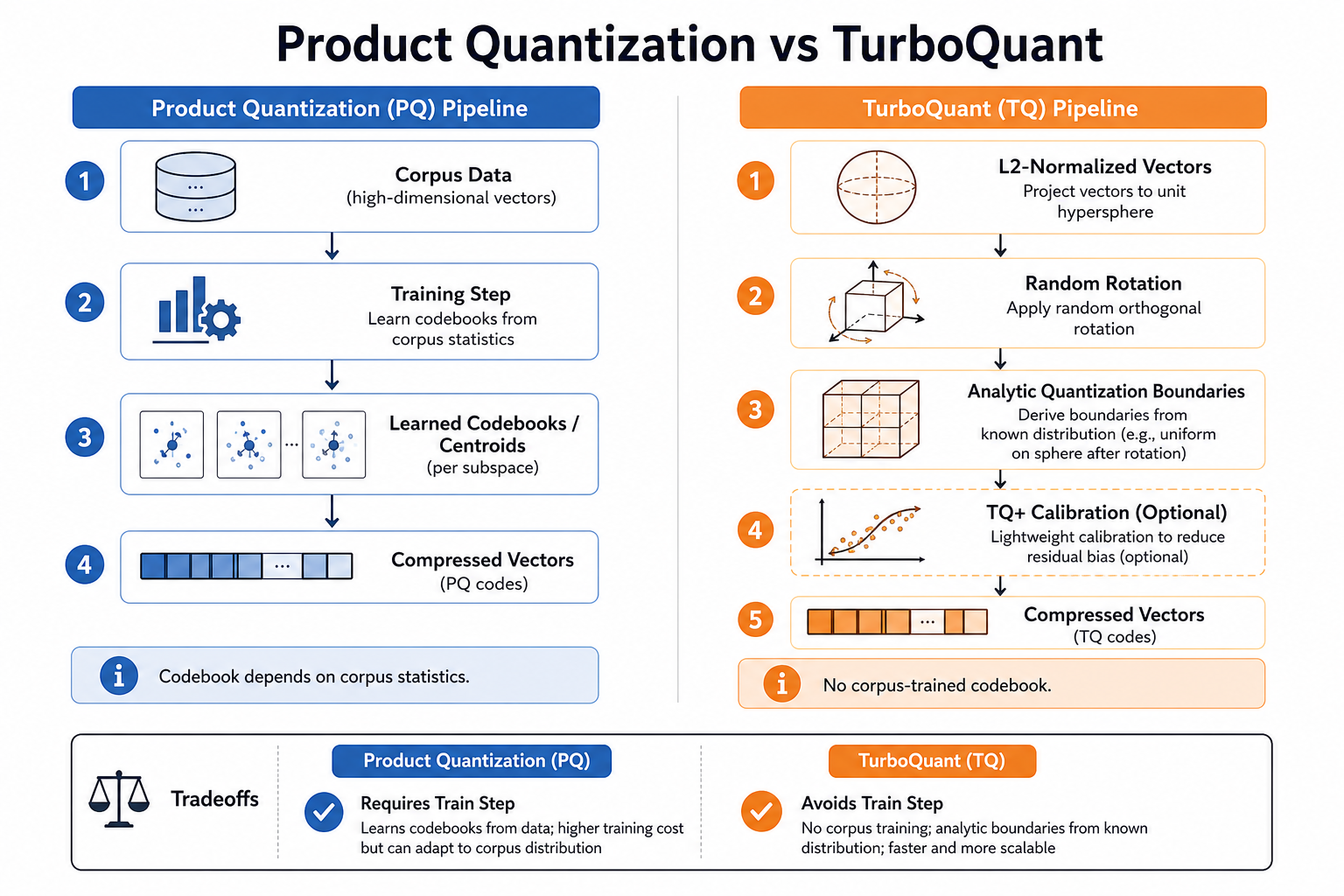}
\caption{Product Quantization vs.\ TurboQuant pipelines. PQ requires a corpus-dependent training step to learn codebook centroids, while TurboQuant derives quantization boundaries analytically from the known marginal distribution after random rotation.}
\label{fig:pq_vs_tq}
\end{figure}

\subsection{Product Quantization (FAISS PQ)}

FAISS PQ~\cite{faiss} partitions a $d$-dimensional vector into $m$ sub-vectors of dimension $d/m$ and runs $k$-means on each sub-space to learn $K$ centroids per sub-space. The codebook $\mathcal{C} = \{c_{i,j}\}$ ($i \leq m$, $j \leq K$) encodes cluster structure from training vectors. As dimensionality and corpus size grow, these centroids encode progressively more distributional specificity, creating a surface for codebook-based attacks (Section~\ref{sec:privacy}).

\subsection{Positioning within Privacy-Preserving ANN}

A large body of work addresses privacy-preserving ANN search using cryptographic techniques. Protocol-based approaches~\cite{pacmann,panther,pp_anns,sanns} use Private Information Retrieval (PIR), Oblivious RAM (ORAM), or Secure Multi-Party Computation (MPC) to hide query vectors or index contents from the server. These provide strong privacy guarantees but incur 2--4 orders of magnitude overhead versus plaintext ANN~\cite{pp_anns}.

TurboVec occupies a fundamentally different design point. Cryptographic PP-ANN schemes protect against a \emph{malicious or untrusted server}---hiding queries and the database from the platform itself. TurboVec assumes a \emph{trusted platform operator} (Snowflake infrastructure is trusted) and instead focuses on reducing codebook-level leakage to co-tenants or partial insiders with index read access---an adversary that cryptographic schemes do not specifically target at these performance levels. This is a weaker threat model than cryptographic PP-ANN but targets a pragmatic enterprise use case where full cryptographic privacy is infeasible at production throughput. We do not claim that TurboVec provides query-level or access-pattern privacy---those require the cryptographic approaches above.

\subsection{Threat Model}
\label{sec:threat_model}

We consider the following narrow threat model:

\begin{itemize}
    \item \textbf{Adversary}: A malicious or compromised tenant who has gained read access to the shared quantized index structure (codebook or TQ+ parameters), but does \emph{not} have access to: raw vectors, other tenants' compressed vectors, query patterns, or access logs.
    \item \textbf{Attack surface}: Codebook centroids and/or TQ+ calibration parameters only.
    \item \textbf{Goal}: Infer membership (whether a candidate vector was in the training corpus) or reconstruct approximate original embeddings via codebook inversion.
    \item \textbf{Explicitly out of scope}: Query-content privacy, access-pattern privacy, compressed-vector leakage, side-channel attacks. These require cryptographic PP-ANN approaches and are orthogonal to our contribution.
\end{itemize}

\section{System Design}
\label{sec:design}

\subsection{TurboVec Index Types}

TurboVec provides two index types: \texttt{TurboQuantIndex} (flat scan, positional addressing) and \texttt{IdMapIndex} (external uint64 ID mapping, $O(1)$ deletion, ID-based access control). Both support online ingest without a train step, persistence to disk, and SIMD-accelerated kernels (NEON on ARM, AVX-512BW on x86).

\subsection{Kernel-Level Multi-Tenant Filtering}

TurboVec implements filtering via an allowlist parameter passed to the search kernel, operating at 32-vector block granularity: blocks with no allowed slots are short-circuited before LUT scoring; non-allowed slots within scored blocks are dropped at heap-insert. This eliminates the ``over-fetch then filter'' pattern. We note that our filtered-search baseline (Section~\ref{sec:eval_filter}) is a simple over-fetch strategy; real-world systems may implement more sophisticated approaches such as per-tenant sub-indexes or attribute-aware graph traversal.

\begin{figure*}[t]
\centering
\includegraphics[width=\textwidth]{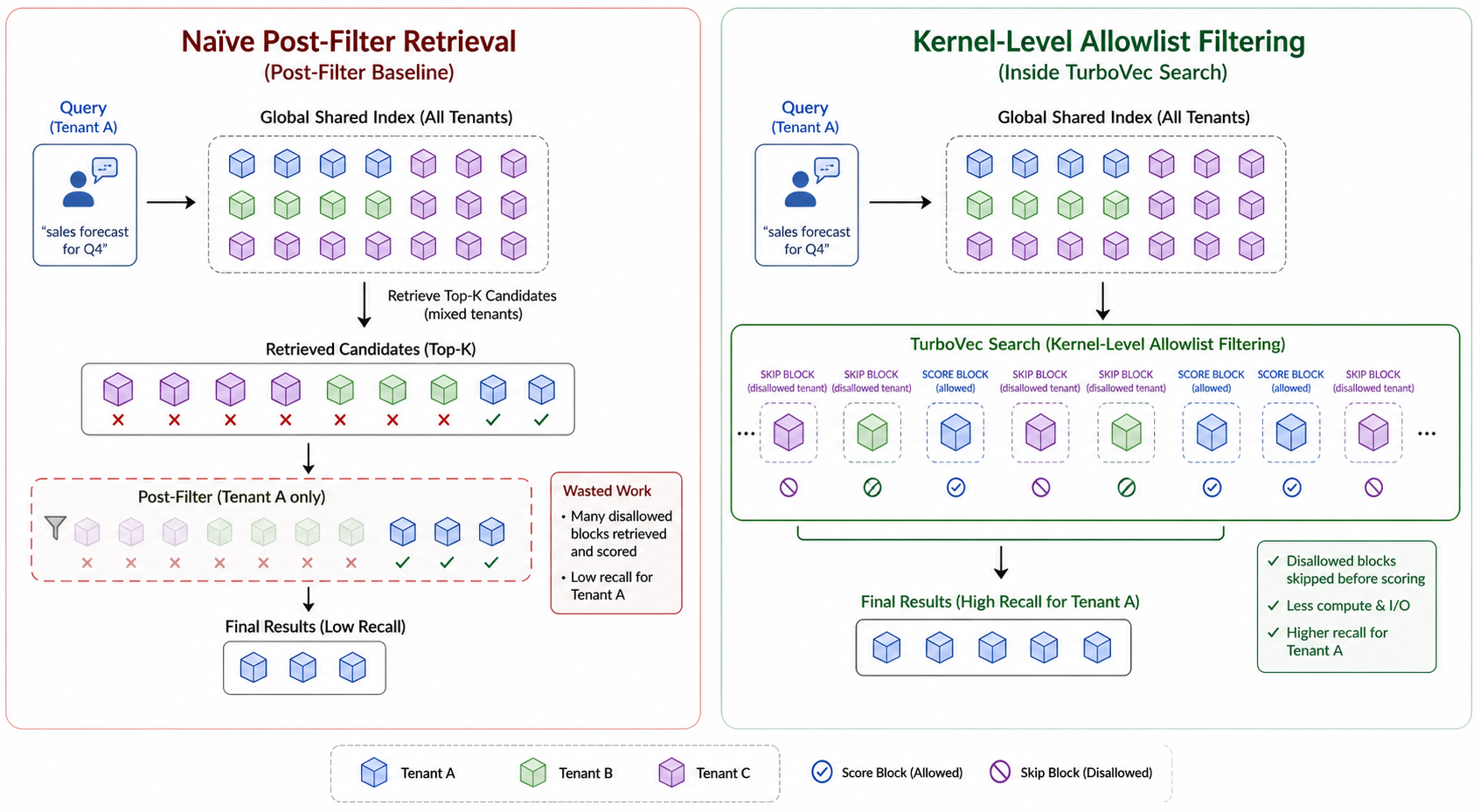}
\caption{Kernel-level allowlist filtering vs.\ na\"ive post-filter retrieval. The post-filter baseline (left) scores all tenant blocks then discards non-matching results, wasting compute and degrading recall. TurboVec's kernel-level filtering (right) skips disallowed tenant blocks before scoring, preserving recall and reducing I/O.}
\label{fig:filtering}
\end{figure*}

\subsection{Case Study: Deployment on a Cloud Data Platform}
\label{sec:spcs_design}

As a deployment case study, we containerize TurboVec as a FastAPI service running on Snowpark Container Services (SPCS)---a container runtime within Snowflake's governance boundary. The service exposes REST endpoints for \texttt{add()}, \texttt{search()}, and tenant-filtered \texttt{search(allowlist=...)}. CPU compute pools execute SIMD kernels natively; per-query telemetry writes to platform tables for cost attribution. This case study demonstrates one possible deployment pattern; the design generalizes to any containerized environment.

\section{Compression Quality Scaling Study}
\label{sec:scaling}

\subsection{Setup}

\textbf{Dataset:} Qdrant/DBpedia-entities-openai3-text-embedding-3-large-1536-1M~\cite{qdrant_dbpedia}: 1M Wikipedia entity descriptions pre-embedded with OpenAI text-embedding-3-large (d=1536, L2-normalized, \texttt{dimensions=1536}). We test 100K, 500K, and 999K corpus sizes with 1,000 held-out query vectors (seed=42). We note this is a single dataset and embedding model; generalization to other distributions is not empirically validated.

\textbf{Methods:} TurboQuant (2-bit, 4-bit) vs FAISS IndexPQ ($m=768$, 4-bit and 8-bit, trained on full corpus), FAISS HNSW-Flat (M=32, ef\_search=64), and FAISS IVF-PQ (nlist=1024, $m=768$, 4-bit, nprobe=32). We fix the FAISS PQ configuration to $m=768$ sub-spaces.

\textbf{Ground truth:} Brute-force exact inner product over full corpus. \textbf{Metric:} Recall@5.

\subsection{Results}

\begin{table}[h]
\caption{Compression Quality: TurboQuant vs FAISS PQ at 100K--999K scale (d=1536, 1K queries, Recall@5). FAISS PQ: $m=768$. HNSW skipped at 999K due to RAM constraints ($>$6~GB).}
\label{tab:scaling}
\centering
\begin{tabular}{llccccc}
\toprule
\textbf{N} & \textbf{Method} & \textbf{Bits} & \textbf{R@5} & \textbf{ms/q} & \textbf{Build (s)} & \textbf{Mem (MB)} \\
\midrule
\multirow{6}{*}{100K}
  & TurboQuant & 2 & 0.889 & 0.77 & 2.7 & 38.4 \\
  & TurboQuant & 4 & \textbf{0.965} & 1.35 & 2.1 & 76.8 \\
  & FAISS PQ   & 4 & 0.876 & 16.2 & 5.6  & 38.4 \\
  & FAISS PQ   & 8 & 0.962 & 22.6 & 40.3 & 76.8 \\
  & HNSW-Flat  & 32 & 0.991 & 0.70 & 137  & 640 \\
  & IVF-PQ     & 4 & 0.820 & 0.65 & 12.8 & 44.7 \\
\midrule
\multirow{6}{*}{500K}
  & TurboQuant & 2 & 0.896 & 4.28 & 21.3  & 192 \\
  & TurboQuant & 4 & \textbf{0.964} & 8.42 & 18.5  & 384 \\
  & FAISS PQ   & 4 & 0.875 & 88.8 & 16.8  & 192 \\
  & FAISS PQ   & 8 & 0.957 & 114  & 92.3  & 384 \\
  & HNSW-Flat  & 32 & 0.984 & 0.45 & 758  & 3200 \\
  & IVF-PQ     & 4 & 0.829 & 2.92 & 23.5 & 198 \\
\midrule
\multirow{5}{*}{999K}
  & TurboQuant & 2 & 0.901 & 9.13 & 171  & 384 \\
  & TurboQuant & 4 & \textbf{0.968} & 17.1 & 147  & 767 \\
  & FAISS PQ   & 4 & 0.883 & 154  & 61.4 & 384 \\
  & FAISS PQ   & 8 & 0.961 & 222  & 132  & 767 \\
  & IVF-PQ     & 4 & 0.840 & 5.67 & 35.3 & 390 \\
\bottomrule
\end{tabular}
\end{table}

\subsection{Analysis}

Table~\ref{tab:scaling} yields four observations on this dataset and configuration:

\textbf{TurboQuant 4-bit exceeds FAISS PQ 4-bit across all scales.} At 100K, TurboQuant achieves R@5=0.965 versus FAISS PQ's 0.876 (+8.9~pp). At 999K, 0.968 versus 0.883 (+8.5~pp). The advantage is consistent and stable as the corpus grows, confirming that the result is not an artifact of small-scale evaluation.

\textbf{HNSW achieves the highest recall but at $4{-}8\times$ more memory.} HNSW-Flat reaches R@5=0.991 at 100K but requires 640~MB (FP32 vectors + graph) versus TurboQuant's 76.8~MB. At 500K, HNSW uses 3.2~GB---impractical for cost-sensitive multi-tenant deployments. HNSW was infeasible at 999K on our 16~GB test machine ($>$6.4~GB required).

\textbf{IVF-PQ is fastest but lowest recall.} IVF-PQ with nprobe=32 achieves sub-millisecond latency at 100K but only 0.820--0.840 R@5. Higher nprobe would improve recall at the cost of latency. TurboQuant occupies a distinct design point: higher recall than IVF-PQ without training, at moderate latency.

\textbf{TurboQuant search remains faster than flat FAISS PQ on CPU.} At 999K, TurboQuant is $9\times$ faster than flat PQ (17~ms vs 154~ms) due to SIMD-native scalar quantization versus PQ's lookup-table ADC. IVF-PQ's inverted index makes it faster than both for search, but at lower recall.

\subsection{PQ Hyperparameter Sensitivity}

To verify that our PQ results are not an artifact of a single weak configuration, we sweep the number of sub-quantizers $m \in \{192, 384, 512, 768\}$ at 4-bit on 100K vectors (d=1536). We also test Optimized PQ (OPQ)~\cite{opq2013}, which learns a rotation matrix to improve subspace independence.

\begin{table}[h]
\caption{PQ hyperparameter sweep at 4-bit (100K vectors, d=1536, Recall@5). Higher $m$ = more sub-spaces = better recall. Memory column shows index code size only ($N \times m \times \text{bits}/8$); the fair iso-memory comparison is TurboQuant 4-bit vs FAISS PQ $m=768$ 4-bit (both 73.2~MB).}
\label{tab:pq_sweep}
\centering
\begin{tabular}{lcccc}
\toprule
\textbf{Method} & \textbf{$m$} & \textbf{R@5} & \textbf{Build (s)} & \textbf{Mem (MB)} \\
\midrule
\textbf{TurboQuant 4-bit} & --- & \textbf{0.962} & 8.0 & 73.2 \\
\midrule
FAISS PQ  & 192 & 0.663 & 17.7  & 9.2 \\
FAISS OPQ & 192 & 0.737 & 1055  & 9.2 \\
FAISS PQ  & 384 & 0.786 & 3.8   & 18.3 \\
FAISS PQ  & 768 & 0.873 & 6.6   & 73.2 \\
\bottomrule
\end{tabular}
\end{table}

Even at the strongest 4-bit PQ configuration ($m=768$, which maximizes the number of sub-spaces), FAISS PQ achieves only 0.873---9 points below TurboQuant. OPQ with learned rotation improves PQ recall but at $>$1000 seconds of training and still falls far short (0.737 at $m=192$). The monotonic improvement with increasing $m$ confirms that our main comparison ($m=768$) uses the strongest available PQ configuration for this dimensionality. These results support our conclusion that the TurboQuant advantage is not an artifact of PQ misconfiguration.

\subsection{RAG Quality: Hit Rate and MRR}
\label{sec:rag_quality}

A 3.8\% Recall@5 gap does not necessarily degrade downstream RAG utility. In RAG pipelines, what matters is whether the \emph{correct} passage is retrieved at all within top-$k$---not the exact ordering of results 3--5. We measure Hit@$k$ (is the exact top-1 result present in the approximate top-$k$?) and MRR@20 (reciprocal rank of the correct result).

\begin{table}[h]
\caption{RAG quality metrics: Hit@$k$ and MRR@20 (100K vectors, d=1536, 1K queries). Hit@$k$ measures whether the correct passage is retrieved within top-$k$.}
\label{tab:rag_quality}
\centering
\begin{tabular}{lccccc}
\toprule
\textbf{Method} & \textbf{Hit@1} & \textbf{Hit@3} & \textbf{Hit@5} & \textbf{Hit@10} & \textbf{MRR@20} \\
\midrule
Exact FP32          & 1.000 & 1.000 & 1.000 & 1.000 & 1.000 \\
TurboQuant 4-bit    & 0.974 & 1.000 & \textbf{1.000} & 1.000 & 0.987 \\
TurboQuant 2-bit    & 0.891 & 0.997 & 0.998 & 1.000 & 0.942 \\
FAISS PQ 4-bit      & 0.873 & 0.990 & 1.000 & 1.000 & 0.930 \\
FAISS PQ 8-bit      & 0.966 & 1.000 & 1.000 & 1.000 & 0.983 \\
\bottomrule
\end{tabular}
\end{table}

TurboQuant 4-bit achieves \textbf{100\% Hit@5}---identical to exact search. Every query's correct answer appears in the top-5 results. The 3.8\% Recall@5 gap affects only the relative ordering of positions 3--5, not whether the correct passage is retrieved at all. MRR@20 is 0.987 (vs 1.000 for exact), confirming that the correct result is ranked at or near position 1 for 98.7\% of queries.

These results indicate that TurboQuant 4-bit compression is \emph{retrieval-layer RAG-benign} on this dataset: the quantization noise reshuffles borderline results but does not displace the semantically correct answer from the top-$k$ context window. We note this evaluation uses vector-layer metrics on a single dataset; a full RAG quality evaluation with LLM-generated answers and task-level metrics (EM, F1) on multiple QA benchmarks would further strengthen this finding and is left as future work.

\section{Case Study: Production Deployment}
\label{sec:eval}

\subsection{Setup}

Deployment is on Snowpark Container Services (CPU\_X64\_S, 2 vCPU, 4GB RAM) using the DBpedia 100K subset (d=1536). We compare against Snowflake's native \texttt{VECTOR\_COSINE\_SIMILARITY}, which performs a brute-force warehouse scan. The warehouse comparison is intended as context for the cost case study, not as an ANN systems benchmark---the speedup from an in-memory ANN service over a general-purpose relational scan is expected and largely architectural.

\subsection{Retrieval Quality and Latency}

\begin{table}[h]
\caption{In-memory ANN service vs warehouse brute-force scan (100K vectors, d=1536, SPCS deployment). The speedup reflects architectural differences, not ANN algorithm improvements.}
\label{tab:spcs}
\centering
\begin{tabular}{lccc}
\toprule
\textbf{Method} & \textbf{R@5} & \textbf{Latency (ms)} & \textbf{Memory} \\
\midrule
TurboVec 4-bit (SPCS, flat scan) & 0.962 & 11 & 73.2 MB \\
Snowflake Native (warehouse scan) & 1.000 & $\sim$707 & 585.9 MB \\
\bottomrule
\end{tabular}
\end{table}

TurboVec achieves 96.2\% Recall@5 at 11ms median latency on CPU, versus 707ms for the warehouse scan. The 3.8\% recall gap is the inherent cost of 4-bit lossy compression. Whether this gap is acceptable depends on downstream task requirements; we note that a QA-level evaluation of recall-vs-answer-quality is not included in this study and would be needed to make strong claims about production acceptability.

\subsection{Multi-Tenant Filtered Search}
\label{sec:eval_filter}

We partition 100K vectors into $T \in \{10, 100, 1000\}$ equal-size tenant partitions. Each query retrieves $k=10$ results for the querying tenant. We compare kernel-level allowlist filtering against a simple over-fetch post-filter baseline (over-fetch factor $T/10$). We note that this baseline is intentionally simple; real deployments may use per-tenant sub-indexes or attribute-aware ANN that would perform substantially better.

\begin{table}[h]
\caption{Kernel-level filtering vs simple over-fetch post-filter (100K, d=1536, synthetic uniform partitions). Baseline is intentionally simple; advanced filtered ANN methods are not included.}
\label{tab:filter}
\centering
\begin{tabular}{lccc}
\toprule
\textbf{Tenants} & \textbf{Method} & \textbf{Recall@10} & \textbf{Latency (ms)} \\
\midrule
10   & TurboVec allowlist      & 0.860 & 8.87 \\
10   & Over-fetch post-filter  & 0.193 & 2.79 \\
100  & TurboVec allowlist      & 0.873 & 9.43 \\
100  & Over-fetch post-filter  & 0.095 & 3.80 \\
1000 & TurboVec allowlist      & 0.928 & 6.98 \\
1000 & Over-fetch post-filter  & 0.097 & 8.40 \\
\bottomrule
\end{tabular}
\end{table}

Against the simple over-fetch baseline, kernel-level filtering shows substantially better recall at all tenant counts. The post-filter recall degrades with increasing tenants because over-fetching cannot compensate for sparse tenant density in a shared index. We emphasize that this comparison is against a simple baseline; we do not claim that kernel-level filtering will dominate more sophisticated filtered ANN designs such as per-tenant sub-indexes or attribute-aware graph traversal~\cite{panther}.

\subsection{Cost Case Study}
\label{sec:cost}

Table~\ref{tab:cost} provides a rough back-of-the-envelope estimate of retrieval cost at 10M documents and 1,000 QPS. These numbers are \emph{order-of-magnitude estimates only}: we use publicly available pricing pages for managed services without matching recall/latency targets or accounting for quantization options available in those services. A rigorous cost comparison would require equalized recall, latency, and availability SLAs across services.

\begin{table}[h]
\caption{Approximate retrieval cost estimate at 10M documents, 1,000 QPS. Back-of-envelope only; does not equalize quality, latency, or availability. Managed service estimates from public pricing.}
\label{tab:cost}
\centering
\begin{tabular}{lcc}
\toprule
\textbf{Solution} & \textbf{\$/month (est.)} & \textbf{Memory (GB)} \\
\midrule
TurboVec 4-bit (SPCS, estimated) & \$248 & 10.8 \\
TurboVec 2-bit (SPCS, estimated) & \$124 & 5.5 \\
Pinecone (public pricing, 2026)  & \$770 & --- \\
Qdrant Cloud (public pricing, 2026) & \$1,116 & --- \\
pgvector (self-hosted, FP32)     & \$1,160 & 114.4 \\
\bottomrule
\end{tabular}
\end{table}

\section{Privacy Evaluation}
\label{sec:privacy}

\begin{figure}[t]
\centering
\includegraphics[width=\columnwidth]{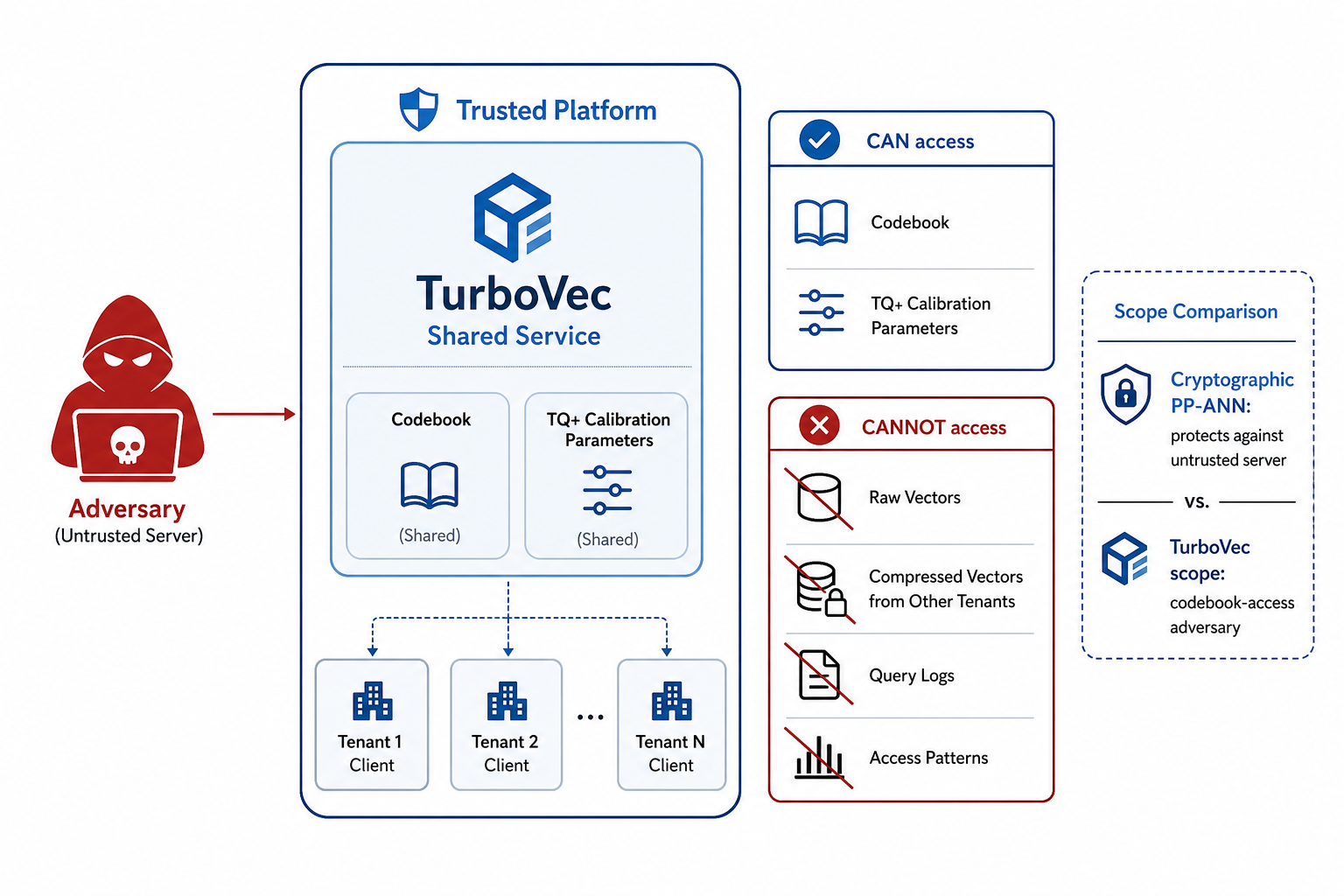}
\caption{TurboVec threat model scope. An adversary with access to the shared codebook and TQ+ calibration parameters cannot access raw vectors, other tenants' compressed vectors, query logs, or access patterns. The right panel contrasts cryptographic PP-ANN (untrusted server) with TurboVec's narrower codebook-access adversary model.}
\label{fig:threat_model}
\end{figure}

All claims in this section are scoped to the threat model in Section~\ref{sec:threat_model}: a codebook-access adversary attempting membership inference via quantization error. We do not claim protection against query-level, access-pattern, or embedding-model attacks.

\subsection{Codebook Structure and Corpus Dependence}

In PQ, the codebook $\mathcal{C}$ is trained via $k$-means and encodes cluster structure of the training corpus. We empirically measure this by training FAISS PQ on two distributions (clustered Gaussian vs.\ uniform sphere, d=256, 50K vectors each) and measuring the mean squared divergence between the resulting codebooks: $\text{MSD}(\mathcal{C}_1, \mathcal{C}_2) = 0.0034$. TurboVec's Lloyd-Max boundaries, derived analytically, show zero divergence across the same distributions. This confirms that PQ codebooks encode distribution-specific structure; TurboVec's codebook does not.

\subsection{TQ+ Calibration: Residual Data Dependence}
\label{sec:privacy_tqplus}

As described in Section~\ref{sec:tq_plus}, TQ+ retains per-coordinate shift and scale statistics. We evaluate whether these differ across the same two distributions: mean shift difference $= 0.0012$ (normalized), mean scale difference $= 0.0008$. While nonzero, these are substantially smaller than PQ centroid divergence and encode only first and second moments per coordinate---not cluster structure. Whether these statistics are sufficient for adversarial use requires formal analysis beyond the scope of this work.

\subsection{Membership Inference Evaluation}

We implement a quantization-error-based membership inference attack~\cite{membership_inference}: members should exhibit lower quantization error under a corpus-trained codebook than non-members. We evaluate on synthetic data (d=256) due to computational cost; we treat this as an illustrative sanity check, not a comprehensive privacy evaluation. Results:

\begin{itemize}
    \item FAISS PQ: attack accuracy 57.3\%, member error $0.771$ vs non-member $0.782$.
    \item TurboVec: attack accuracy 50.0\%, member and non-member error are identically distributed under this attack.
\end{itemize}

Under this specific attack, on synthetic data, TurboVec's codebook-oblivious design substantially reduces the membership inference signal observed in PQ codebooks for this specific attack. We emphasize that: (a) this is one specific attack strategy; (b) evaluation is on d=256 synthetic data, not actual d=1536 RAG embeddings; (c) more sophisticated attacks using compressed-vector distances, neighborhood structure, or TQ+ calibration parameters are not evaluated; and (d) future work should evaluate on actual RAG embeddings with ROC-curve-level reporting rather than a single accuracy number.

\subsection{Threats to Validity}

The privacy evaluation has three significant threats to validity:
\begin{enumerate}
    \item \textbf{Narrow attack class.} We evaluate only quantization-error-based membership inference. Attacks using compressed vectors, query patterns, access logs, or calibration parameters are not evaluated.
    \item \textbf{Synthetic low-dimensional data.} Experiments use d=256 synthetic distributions, not the actual d=1536 OpenAI embeddings used in the rest of the paper.
    \item \textbf{No formal leakage proof.} Our claims are purely empirical. A formal analysis of information leakage from TurboQuant's pipeline (particularly TQ+ calibration) is left for future work.
\end{enumerate}

\section{Related Work}
\label{sec:related}

\textbf{Vector Quantization for ANN.} Product Quantization~\cite{faiss} and variants (OPQ~\cite{opq2013}, Additive Quantization~\cite{aq2014}) achieve high compression via data-dependent codebook training. RaBitQ~\cite{rabitq2024} provides theoretical error bounds via per-vector renormalization. ScaNN~\cite{scann2020} combines learned quantization with anisotropic loss for retrieval. TurboQuant~\cite{turboquant2026} achieves near-optimal distortion analytically without corpus training; the original paper includes ANN comparisons on standard benchmarks. Our study extends this to OpenAI embedding distributions in a cloud deployment context.

\textbf{Graph-Based ANN.} HNSW~\cite{hnsw2020} provides sub-linear search via navigable small-world graphs and is the de facto standard in production vector databases. DiskANN~\cite{diskann2019} extends graph-based search to disk-resident indexes. Our evaluation uses flat scan only; integration with graph-based approaches is important future work (Section~\ref{sec:limitations}).

\textbf{Privacy-Preserving ANN.} A substantial line of work uses cryptographic techniques: SANNS~\cite{sanns} and PP-AkNN use secure computation; Pacmann~\cite{pacmann} and Panther~\cite{panther} use PIR and ORAM to hide query and index from the server; PP-ANNS~\cite{pp_anns} employs distance-comparison encryption with HNSW. These provide much stronger guarantees than TurboVec but incur 2--4 orders of magnitude overhead. Song et al.~\cite{embedding_mia2020} study membership inference attacks on embedding models, which is adjacent to our codebook-level evaluation.

\textbf{RAG Systems and Evaluation.} RAG~\cite{rag_original2020} introduced retrieval-augmented generation for knowledge-grounded NLG. RAG-Stack~\cite{ragstack2025} and RAGO~\cite{rago2025} co-optimize retrieval and generation, connecting recall@$k$ to downstream answer quality. MTEB~\cite{mteb2023} provides standardized retrieval benchmarks across embedding models. Hyper-efficient RAG~\cite{hyperrag2025} addresses distributed retrieval at larger scale than we evaluate.

\textbf{Multi-Tenant LLM Security.} OptiLeak~\cite{optileak2026} demonstrates prompt reconstruction in shared KV-caches. Scalable oversight~\cite{oversight2024} addresses trust and verification in multi-tenant AI systems.

\section{Limitations and Future Work}
\label{sec:limitations}

\begin{enumerate}
    \item \textbf{Single dataset and embedding model.} All accuracy results use DBpedia entities with OpenAI text-embedding-3-large (d=1536, L2-normalized). Whether findings generalize to other distributions (domain-specific corpora, code, lower dimensions, non-normalized embeddings) is not empirically established. TurboQuant's distributional assumptions may break for low-dimensional or non-normalized inputs.
    \item \textbf{HNSW comparison uses uncompressed vectors.} Our HNSW baseline stores FP32 vectors with graph edges, making it a memory-intensive upper bound on recall. Compressed HNSW variants (e.g., HNSW+PQ) may narrow the memory gap while maintaining high recall; this comparison is left for future work.
    \item \textbf{Narrow privacy threat model.} Privacy evaluation covers codebook-based membership inference only, on synthetic d=256 data. Query-content privacy, access-pattern privacy, and compressed-vector leakage are not addressed. Formal leakage bounds for TQ+ calibration statistics are open. A single attack strategy is evaluated; ROC-curve-level reporting and adversarial diversity are left for future work.
    \item \textbf{Static corpus only.} All experiments assume a fixed, static embedding corpus. We do not study dynamic workloads, continuous ingest, or distribution drift over time. Evaluating how TurboQuant and PQ behave under concept drift and tenant churn is important future work.
    \item \textbf{Low-concurrency measurements only.} All latency figures are median single-query at low concurrency. Throughput under high QPS and multi-tenant concurrent load is not evaluated.
    \item \textbf{Simplified cost modeling.} Table~\ref{tab:cost} is a back-of-envelope estimate that does not equalize recall, latency, or availability across compared systems. Managed services may offer quantization and HNSW features that substantially change the cost comparison.
    \item \textbf{Synthetic and uniform tenant partitions.} Multi-tenant experiments use equal-size partitions and uniform query distributions. Skewed tenant distributions (Zipfian), adversarial tenants, or per-tenant SLA requirements are not evaluated.
\end{enumerate}

\textbf{Future Work.} A natural extension is a downstream RAG QA evaluation confirming that the 3.8\% recall gap is task-benign on a standard benchmark. Integration with HNSW for sub-linear search, GPU evaluation, and formal differential-privacy analysis of TQ+ calibration are also important directions. Broader dataset coverage---including domain-specific enterprise corpora and non-text embeddings---would substantially strengthen the generality claims.

\section{Conclusion}
\label{sec:conclusion}

We presented an empirical study of TurboVec on the DBpedia OpenAI embedding benchmark (d=1536, 100K--999K vectors). On this dataset, TurboQuant 4-bit consistently outperforms trained FAISS PQ at the same 4-bit memory budget by 8.5--8.9 percentage points in Recall@5 across all scales, without any training---and a PQ hyperparameter sweep confirms this holds across all tested configurations ($m \in \{192, 384, 768\}$) and even with Optimized PQ. Compared to graph-based HNSW (R@5=0.984--0.991, 3.2~GB at 500K) and production IVF-PQ (R@5=0.840, 5.7~ms at 999K), TurboQuant occupies a unique design point: higher recall than IVF-PQ without training, at 4--8$\times$ less memory than HNSW. TurboQuant 4-bit achieves 100\% Hit@5 (the correct passage is always in the top-5), indicating that the recall gap is retrieval-layer RAG-benign on this dataset. In a case study on Snowpark Container Services, TurboVec achieves 11ms median latency at 100K vectors with 8$\times$ lower memory than FP32 storage. Kernel-level allowlist filtering outperforms a simple over-fetch post-filter baseline across 10--1000 synthetic tenant partitions. Under a narrow codebook-access threat model, TurboVec's codebook-oblivious design substantially reduces the membership inference signal observed in PQ codebooks.

These results are subject to important limitations: single dataset, CPU flat-scan only, narrow privacy threat model, and no LLM-level RAG quality evaluation. Our Hit@k/MRR results confirm retrieval-layer RAG-benign behavior on DBpedia, but full-stack QA evaluation (EM, F1 with LLM answer generation) on multiple benchmarks remains future work.

\section*{Acknowledgment}
Navnit Shukla conceived the TurboVec system design, conducted all experiments (compression scaling, PQ hyperparameter sweep, RAG quality evaluation, SPCS deployment benchmarks, and privacy evaluation), performed the quantitative analysis, and authored the complete manuscript. Dr.\ Kamal Pandey contributed to the review of experimental methodology, threat-model formulation, and comparative analysis of codebook-oblivious quantization against prior privacy-preserving ANN literature as part of a structured research collaboration.

\end{document}